%% file: main.tex
\documentclass[fleqn,10pt]{wlscirep}
\usepackage[utf8]{inputenc}
\usepackage{import}
\usepackage[T1]{fontenc}
\usepackage{graphicx}
\usepackage{movie15}
\usepackage{soul}
\usepackage{longtable,booktabs, array} 
\usepackage{multirow}
\usepackage{amsmath}
\usepackage{xhfill}

\title{Challenges in data-based geospatial modeling for environmental research and practice}

\author[1]{Diana Koldasbayeva}
\author[1]{Polina Tregubova}
\author[1]{Mikhail Gasanov}
\author[1,2]{Alexey Zaytsev}
\author[1]{Anna Petrovskaia}
\author[1,3]{Evgeny Burnaev}

\affil[*]{diana.koldasbayeva@skoltech.ru}
\affil[1]{Skolkovo Institute of Science and Technology, Bolshoy Boulevard 30, bld. 1, 121205 Moscow, Russia}
\affil[2]{Yanqi Lake Beijing Institute of Mathematical Sciences and Applications (BIMSA), No. 544, Hefangkou Village, Huaibei Town, Huairou District, Beijing 101408}
\affil[3]{Autonomous Non-Profit Organization Artificial Intelligence Research Institute (AIRI), 105064 Moscow, Russia}

\begin{abstract}

With the rise of electronic data, particularly Earth observation data, data-based geospatial modelling using machine learning (ML) has gained popularity in environmental research. Accurate geospatial predictions are vital for domain research based on ecosystem monitoring and quality assessment and for policy-making and action planning, considering effective management of natural resources. The accuracy and computation speed of ML has generally proved efficient. However, many questions have yet to be addressed to obtain precise and reproducible results suitable for further use in both research and practice. A better understanding of the ML concepts applicable to geospatial problems enhances the development of data science tools providing transparent information crucial for making decisions on global challenges such as biosphere degradation and climate change. This survey reviews common nuances in geospatial modelling, such as imbalanced data, spatial autocorrelation, prediction errors, model generalisation, domain specificity, and uncertainty estimation. We provide an overview of techniques and popular programming tools to overcome or account for the challenges. We also discuss prospects for geospatial Artificial Intelligence in environmental applications.

\end{abstract}
\begin{document}

\flushbottom
\maketitle
%
%

\begin{figure}[!h]
\centering
\includegraphics[width=.7\textwidth]{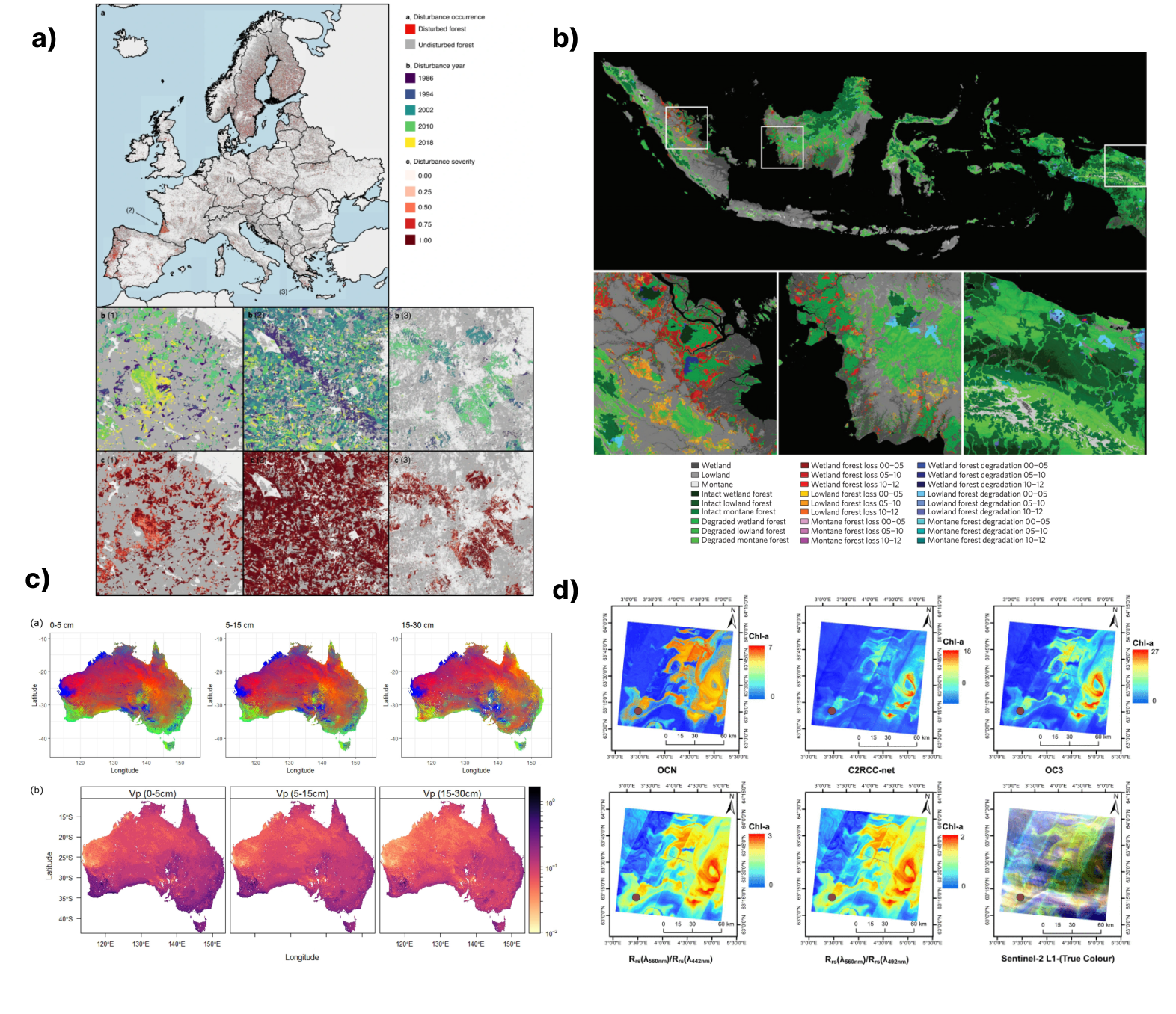}
\caption{Examples of geospatial mapping performed for different tasks of environmental monitoring and assessment a) maps of forest disturbance regimes of Europe \cite{senf2021mapping}; b) land cover and mapping of losses for different types of forest in Indonesia \cite{margono2014primary}; c) maps of soil organic carbon (SOC) fractions contribution to SOC for selected depths of 0–5, 5–15, and 15–30 cm obtained for Australia \cite{roman2023mapping}; d) maps of chlorophyll-A estimation derived from Sentinel-2 data in the Barents Sea \cite{asim2021improving}.}
\label{fig:examples_maps}
\end{figure}

To date, obtaining spatial predictions is an essential step in the monitoring, assessment, and prognosis tasks applicable to all kinds of Earth systems on both local and global scales (Figure \ref{fig:examples_maps}). Regional spatial analysis for areas of interest now plays a crucial role in risk-sensitive land use and vulnerability assessment facing environmental sustainability threats, climate change urgency, and disasters occurrences such as fires\cite{giglio2009active, chuvieco2019historical, mohajane2021application}, floods\cite{uddin2019operational, tarpanelli2022effectiveness, tavus2022flood}, and droughts \cite{hoque2020assessing, lu2020mapping}, in biodiversity conservation prioritisation and actions planning  \cite{verstegen2019recent, jetz2019essential, moilanen2020practical}, natural resources inventorying \cite{zuo2019deep, tapia2021much, heinrich2023carbon}, land cover inventorying and change detection \cite{karra2021global, brown2022dynamic}, ecosystems functioning assessment \cite{yang2019mapping, orsi2020mapping}, and other environment-related tasks. 

Spatial modelling results could be not only the final expected outcome but an intermediate step and required base for the following system analysis. For instance, forest maps can be used to estimate how vulnerable vegetation is to events contributing to climate change, such as cycles of forest damage and forest succession after fires \cite{bouchard2008fire, syphard2018mapping}, and to assess the long-term sustainability of forest carbon sinks \cite{fan2023siberian}. Another example is a prediction of the quality of resources such as soil \cite{schillaci2017spatio} based on environmental predictor maps. One of the most common use cases is applying land use and land cover (LULC) map products for a wide range of research and practical issues. 
Land cover maps can be used to estimate environment-related phenomena, such as ecosystem services \cite{yang2019mapping, orsi2020mapping}, assess spatiotemporal resource changes, and distinguish influencing factors \cite{schillaci2017spatio, keskin2019digital}. Apart from that, LULC products serve to enhance prediction — for example, to stratify modelling solutions (ensembling) in order to raise forecast precision \cite{bjaanes2021deep}. The products can also serve as label data to develop new prediction approaches—for example, to classify single-date images in order to obtain large area cover maps \cite{zhang2023demonstration}.

The expectations about mapping usefulness for developing decision-making tools have been quite high since at least the beginning of the century\cite{gewin2004mapping}. Being not only a tool for purely increasing our knowledge about the environment, geospatial predictions have already been included as an essential base for policy and coordinated action support. For instance, fire mapping supports The Monitoring Trends in Burn Severity (MTBS) program \cite{eidenshink2007project}, catching burn severity and extent of large fires for monitoring the effectiveness of the National Fire Plan. Invasive species habitat suitability mapping informs decision-making by identifying high-risk species and pathways, increasing information exchange, action efficiency, and cost-savings within the U.S. Department of the Interior Invasive Species Strategic Plan \cite{strategic2021plan}. Another example is the geospatial assessment and management of flood risks as an information tool to plan and prioritise technical, financial, and political decisions regarding flood risk management within Directive 2007/60/EC (2007) \cite{ES2007flood}. It is highlighted that Earth observation global maps play a crucial role in supporting the key aspects of the Paris Agreement, such as making nationally determined contributions, enhancing the transparency of national GHG (greenhouse gas) reporting, managing GHG sinks and reservoirs, and developing market-based solutions \cite{melo2023satellite}. 

On a global scale, spatial mapping results can serve as both inputs for integrated assessment models (IAMs) and target output data to forecast and understand postponed consequences of changing socioeconomic development and climate change scenarios, which helps to plan climate change actions considering other sustainable development goals\cite{rogelj2018mitigation}. Additionally, information from global mapping products can fill the blind spots where domestic land cover inventories are poorly organised and impede coordinating responses to global challenges \cite{melo2023satellite}. 



At the same time, the question of the quality of spatial predictions and possible struggles to achieve trustworthy results has been drawing much attention recently. One of the most important concerns lies in the very nature of data-based modelling—that is, the belief that knowledge can be obtained through observation \cite{janowicz2023philosophical}. Thus, proper techniques for managing data from geospatial observations are of great question. Another issue related to efficient and fair data handling is the existing gap between domain specialists and applied data scientists, both underrepresented in each other’s fields.

In recent work \cite{ploton2020spatial} it was emphasised that ignoring the spatial nature of the data led to the misleading high predictive power of the model, while appropriate spatial model validation methods revealed poor relationships between the target characteristic—aboveground forest biomass—and selected predictors. On the contrary, in  \cite{wadoux2021spatial} the idea of spatial validation is critically discussed, while other approaches to overcome biases in the data are proposed instead. The importance of spatial dependence between training and test sets and its influence on the model generalisation capabilities in the Earth observation data classification is addressed in \cite{karasiak2022spatial}. Other examples of issues in global environmental spatial mapping are the distribution shift, data concentration, and predictions’ accuracy assessment, which are discussed in the latest comment article \cite{meyer2022machine}.
Thus, given the confusion about the modelling process and quality estimation of results and in light of the rising demand for spatial predictions, an overview of common struggles in geospatial modelling and relevant approaches and tools to address the issues are of both scientific and practical use.

\begin{figure}[!h]
\centering
\includegraphics[width=.9\textwidth]{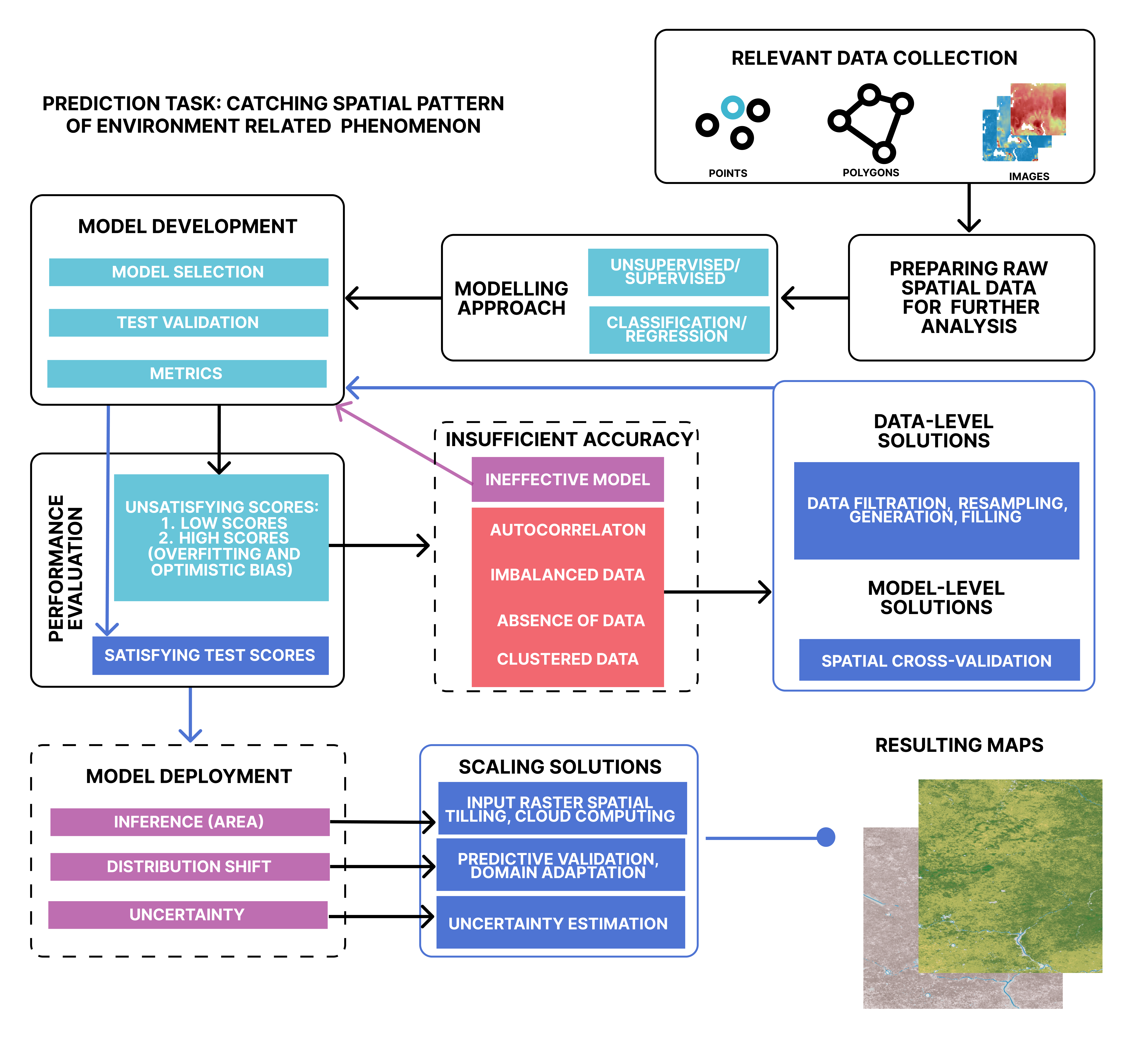}
\caption{General workflow for the tasks including geospatial modelling process and common issues relevant for each stage.}
\label{fig:workflow}
\end{figure}

In addition to the existing literature background \cite{kanevski2009machine, li2011application, 
dale2014spatial, thessen2016adoption, feng2019checklist, meyer2019importance, tahmasebi2020machine, meyer2022machine}, this review aims to comprehensively address the limitations of data-driven geospatial mapping at each step of predicting the spatial distribution of target features. Here we provide a practical guide, discussing the challenges associated with using nonuniformly distributed real-world data from various domains in environmental research, including those from open sources. These challenges include dealing with limited observations and imbalanced and autocorrelated data, maintaining the model training process, and assessing prediction quality and uncertainty (Figure \ref{fig:workflow}). Throughout the review, we provide examples from recent environmental geospatial modelling research to illustrate the identified problems, highlight the underlying theoretical concepts, and present approaches to evaluating and overcoming each specific limitation.

\section{Data-driven approaches to forecasting spatial distribution of environmental features}

In this review, we analyse geospatial modelling based on data-driven approaches, meaning that models are built with parameters learned from observations’ data, thus simulating new data minimally different from the “ground truth” under the same set of descriptive features. Among the standards guiding the implementation of data-based model applications, CRISP-DM is the most well-known. There are, however, other workflows with more nuanced guidelines tailored to specific problems or more mature fields of data-based modelling \cite{azevedo2008kdd, schroer2021systematic}. Recently, guidelines and checklists have been proposed for environmental modelling tasks to help address common problems and improve the reliability of outputs \cite{feng2019checklist, sillero2021want}. For instance, a checklist \cite{feng2019checklist} for ecological niche modelling suggests using a standardised format for reporting the modelling procedure and results to ensure research reproducibility. It emphasises the importance of disclosing details of each prediction-obtaining step, from data collection to model application and result evaluation. In general, the main steps to solve the applied problems using data-driven algorithms can be the following~\cite{wirth2000crisp}: 

\begin{enumerate}
\item Understanding the problem and the data. This step depends on the specific domain, such as conservation biology and ecology, epidemiology, spatial planning, natural resource management, climate monitoring, and predicting hazardous events.
\item Data collection and feature engineering. Pre-processing data from different domains involves collecting ground-truth data from specific locations and combining it with relevant environmental features such as, for instance, Earth observation images, weather and climate patterns.
\item Model selection. The choice of model depends on the characteristics of the target feature, the specificity of the task, and available resources.
\item Model training. Training the model involves optimizing hyperparameters to fit the data type and shape.
\item Accuracy evaluation. Appropriate accuracy scores are selected based on the task, with a focus on controlling overfitting. The model's performance is better to be evaluated using "gold standard" data with expert annotations.
\item Model deployment and inference. This involves building maps with spatial predictions for the region of interest and determining the level of certainty of the model's estimations.
\end{enumerate}

For data-based modelling tasks, including mapping, various classic machine learning (ML) algorithms \cite{wang2019crop} and deep learning (DL) algorithms \cite{wang2019comparison, yuan2020deep, karra2021global, brown2022dynamic} are used. The choice of algorithm depends on the type of target variable. Classification algorithms are employed for predicting categorical target variables, which could be land cover and land change mapping \cite{ karra2021global, brown2022dynamic}, cropland and crop type mapping \cite{wang2019crop, you202110}, identification of pollution sources \cite{jia2019methodological}, mapping pollutant impact to distinguish free and affected lands \cite{ozigis2019mapping}, the landslide \cite{wang2019comparison} and wildfire \cite{bjaanes2021deep} susceptibility mapping, and habitat suitability mapping \cite{hamilton2022increasing}. Regression algorithms are used to forecast the distribution of continuous target variables – for instance, the prediction of the geospatial distribution of important soil features, such as soil carbon characteristics \cite{keskin2019digital}, groundwater potential, and quality assessment \cite{panahi2020spatial, nikitin2021regulation}, and vegetation characteristics such as forest height \cite{potapov2021mapping} and biomass \cite{harris2021global}. Handling data and interpreting results at each step of obtaining spatial predictions can be complex, leading to low-quality predictions and misleading interpretations. Therefore, careful control using adopted approaches and metrics is necessary. 

\section{Imbalanced data}

\subsection{Problem statement}
The problem of imbalanced data is one of the most relevant issues in environment-related research with a focus on spatial capturing of target events or features. Imbalance occurs when the number of samples belonging to one class or classes (majority class[es]) significantly surpasses the number of objects in another class or classes (minority class[es]) \cite{kubat1997addressing, kaur2019systematic}.

Although being highly imbalanced is one of the basic properties of the real world, most models assume uniform data distribution and complete input information. Thus, a nonuniform input data distribution poses difficulties when training models. The minority class occurrences are infrequent, and classification rules that predict the small classes are usually rare, overlooked, or ignored. As a result, test samples belonging to the minority classes are misclassified more frequently compared with test samples from the predominant classes.

In geospatial modelling, one of the most frequent challenges is dealing with sparse or nonexistent data in certain regions or classes \cite{jasiewicz2015geo, langford2018wildfire, shaeri2019application, benkendorf2020effects, sharma2022use}. This issue arises from the high cost of data collection and storage, methodological challenges, or the rarity of certain phenomena in specific regions.

For instance, forecasting habitat suitability for species --- species distribution modelling (SDM) --- is a common task in conservation biology, and it relies on ML methods, often involving binary classification of species abundance. Although well-known sources such as the GBIF (the Global Biodiversity Information Facility) database~\cite{GBIF} provide numerous species occurrence records, absence records are few, while it is additionally difficult to establish such locations from the methodological point of view~\cite{anderson2016final}.
For instance, anomaly detection and mapping, particularly relevant for ecosystem degradation monitoring, often involves the challenge of overcoming imbalanced data—for example, in pollution cases, such as oil spills occurring on both land and water surfaces. Accurate detection and segmentation of oil spills with image analysis is vital for effective leak cleanup and environmental protection. But, despite the regular collection of Earth surface images by various satellite missions, there are significantly fewer scenes of oil spills compared with images of clean water~\cite{kubat1998machine, shaban2021deep}. Similarly, detection and mapping of hazardous events, such as wildfires, struggles from the same problem \cite{langford2018wildfire}. 

In classic research, Weiss and Provost ~\cite{weiss2003learning} examined the relationship between decision trees' classification abilities and the class distribution of training data and demonstrated that a relatively balanced distribution generally yields better results compared with an imbalanced one.
The sample size plays a critical role in assessing the accuracy of a classification model considering the class imbalance. When the imbalance degree remains constant, the limited sample size raises concerns about discovering inherent patterns in the minority class. Experimental findings suggest that the significant error rate caused by imbalanced class distribution decreases as the training set size increases~\cite{japkowicz2002class}. This observation aligns logically, because having more data provides the classification model with a better understanding of the minority class, enabling differentiation between rare samples and the majority. According to Japkowicz~\cite{japkowicz2002class}, if a sufficiently large dataset is available and the training time for such a dataset is acceptable, the imbalanced class distribution may not hinder the construction of an accurate classification model~\cite{sun2009classification}.


\subsection{Approaches to measuring the problem of imbalanced data}
Various approaches quantify class imbalance. One method is to examine the class distribution ratio directly, which can be as extreme as 1:100, 1:1000, or even more in real-world scenarios. The minority class percentage (MCP) calculates the percentage of instances in the minority class. Gini index (GI) measures inequality or impurity among classes, indicating imbalance~\cite{he2009learning}. Shannon entropy (SE) is another way to measure non-uniformity or data substance and can be linked to imbalance through the entropy of the class distribution~\cite{he2009learning}. The Kullback-Leibler (KL) divergence measures the contrast between probability distributions. Thus, it shows how close the observed class distribution is to a hypothetical balanced distribution~\cite{chawla2002smote}. Higher values of GI, SE, and KL indicate a higher imbalance.

In dealing with class imbalance, it is crucial to use appropriate quality metrics to reflect model performance accurately. Standard accuracy may mislead, especially when there is a significant class imbalance --- for example, a model that always predicts the major class yielding a high accuracy but performs poorly for the minority class~\cite{van1979information}. The F1 score, combining precision and recall, is a better alternative and is commonly used for imbalanced data, particularly for the minority class. Another useful metric is the G-mean, which balances sensitivity and specificity and provides a more reliable performance assessment, especially in imbalanced datasets~\cite{japkowicz2011evaluating, japkowicz2002class}.

\subsection{Solutions to improve geospatial modelling for imbalanced data}


Various reviews address imbalanced data in ML, in general, \cite{sun2009classification, he2009learning, krawczyk2016learning, kaur2019systematic}, while approaches relevant to geospatial modelling are also worth to be discussed. Approaches to tackling imbalanced data problems in geospatial prediction tasks can be divided into data-level, model-level and combined techniques.

\subsubsection{Data-level approaches} 

\begin{figure}[!h]
\centering
\includegraphics[width=.9\textwidth]{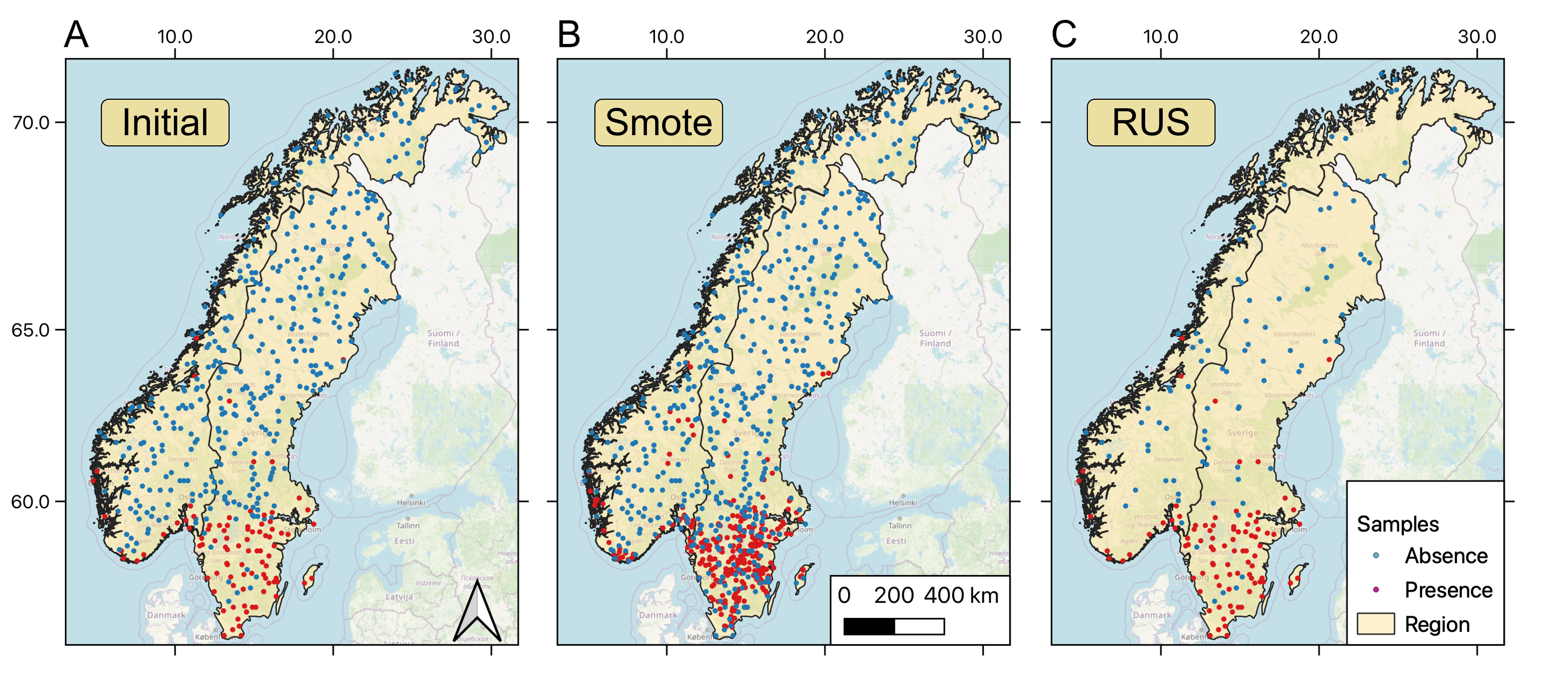}
\caption{Handling imbalance data for artificially species distribution generated data. A) Data generation using \texttt{virtualspecies}~\cite{leroy2016virtualspecies} R package based on annual mean temperature and annual precipitation, obtained from WordlClim~\cite{fick2017worldclim} database. B) Oversampling the minority class by smote method with \texttt{smotefamily}~\cite{smotefamily} R package. C) Achieving a balanced dataset through random undersampling of the prevalent class.}
\label{fig:samples}
\end{figure}

\paragraph{Numerical data}
In terms of working with the data itself, the class imbalance problem can be addressed by modifying the training data through resampling techniques. There are two main ideas: oversampling the minority class and undersampling the majority class~\cite{chawla2002smote, chawla2004special, estabrooks2000combination, sun2009classification}. These techniques can be applied randomly or in an informative way. For instance, in SDM, random oversampling is often a choice to create new minority class samples (e.g., species absence)~\cite{thuiller2009biomod}, while random undersampling is used to balance the class distribution, particularly for species occurrence~\cite{aiello2015spthin}. Informative oversampling may involve generating artificial minority samples based on geographic distance. For instance, in SDM, pseudoabsence generation can be performed using the \texttt{biomod2} R package~\cite{thuiller2009biomod}  with a 'disk' option based on geographic distance. Informative undersampling can involve thinning the majority class by deleting geographically close points, which can be done with \texttt{spThin} R package \cite{aiello2015spthin}. In Figure \ref{fig:samples}, we illustrate the issue of imbalanced data and present solutions, including oversampling and undersampling techniques.

More complex methods for handling imbalanced data involve adding artificial objects to the minority class or modifying samples in a meaningful way. One popular approach is the synthetic minority oversampling technique (SMOTE)~\cite{chawla2002smote}, which combines both oversampling of the minority class and undersampling of the majority class. SMOTE creates new samples by linearly interpolating between minority class samples and their K-nearest neighbour minority class samples.

 

Various has recently seen various modifications \cite{shelke2017review, kovacs2019empirical}. Since there are more than 100 SMOTE variants in total \cite{fernandez2018smote}, here we focus on those relevant to geospatial modelling. One widely used method for oversampling the minority class is the Adaptive synthetic sampling approach for imbalanced learning (ADASYN) \cite{zhang2020seismic, perez2021machine, cao2022evaluating, gomez2022multiclass}. ADASYN uses a weighted distribution that considers the learning difficulties of distinct instances within the minority class, generating more synthetic data for challenging instances and fewer for less challenging ones \cite{he2008adasyn}.To address potential overgeneralization in SMOTE~\cite{he2009learning, shelke2017review}, Borderline-SMOTE is proposed. It concentrates on minority samples that are close to the decision boundary between classes. These samples are considered to be more informative for improving the performance of the classification model on the minority class. Two techniques, Borderline-SMOTE1 and Borderline-SMOTE2, have been proposed, outperforming SMOTE in terms of suitable model performance metrics, such as the true-positive rate and an F-value~\cite{han2005borderline}. Another approach is the Majority Weighted Minority Oversampling Technique (MWMOTE), which assigns weights to hard-to-learn minority class samples based on their Euclidean distance from the nearest majority class samples~\cite{barua2012mwmote}. The algorithm involves three steps: selecting informative minority samples, assigning selection weights, and generating synthetic samples using clustering. MWMOTE consistently outperformed other techniques such as SMOTE, ADASYN, and RAMO in various performance metrics, including accuracy, precision, F-score, G-mean, and the Area Under the Receiver Operating Characteristic (ROC) Curve (AUC)~\cite{barua2012mwmote}.

As for the limitations of discussed data-level approaches, oversampling and undersampling may, given that they are widely used, lead to overfitting and introduce bias in the data~\cite{he2009learning, fernandez2018smote}. Additionally, these techniques do not address the root cause of class imbalance and may not generalise well to unseen data~\cite{batista2004study, sun2009classification}. 

\paragraph{Image data}

Computer vision techniques applied to Earth observation tasks have gained their popularity and now play a pivotal role in the analysis of remote sensing data\cite{shamsolmoali2019novel, nowakowski2021crop, illarionova2022estimation, karra2021global}. Thus, it is worth examining approaches to overcoming data imbalance problems on the image level.

Data augmentation is a fundamental technique for expanding limited image datasets~\cite{simard2003best}. It revolves around enriching training data by applying various transformations, such as geometric alterations, colour adjustments, image blending, kernel filters, and random erasing. These transformations enhance both model performance and generalization.
Geospatial modelling frequently uses data augmentation strategies to address specific challenges. For example, experts employ a cropping-based augmentation approach in mineral prospective mapping. This technique generates additional training samples while preserving the spatial distribution of geological data~\cite{yang2022applications}.
DL-based oversampling techniques such as adversarial training, Neural Style Transfer, Generative Adversarial Networks (GANs) and meta-learning approaches offer intelligent alternatives for oversampling ~\cite{khosla2020enhancing}. 
Neural Style Transfer stands out as a captivating method for generating novel images. It achieves this by extrapolating styles from external sources or blending styles among dataset instances~\cite{gatys2015neural}. For instance, researchers have harnessed the power of Neural Style Transfer alongside ship simulation samples in remote sensing ship image classification. This dynamic combination enhances training data diversity, resulting in substantial improvements in classification performance~\cite{xiao2021progressive}.
GANs, on the other hand, specialise in crafting artificial samples that closely mimic the characteristics of the original dataset. For instance, GANs have been used for data augmentation in specific domains, such as roof damage detection and partial discharge pattern recognition in Geographic Information Systems~\cite{asami2022data, wang2022gan}. In the context of landslide susceptibility mapping, a notable research study introduces a GAN-based approach to tackle imbalanced data challenges, comparing its effectiveness with traditional methods such as SMOTE~\cite{al2021new}.


Taking it a step further, researchers have unveiled a deeply supervised Generative Adversarial Network (D-sGAN) tailored for high-quality data augmentation of remote sensing images. This innovative approach proves particularly beneficial for semantic interpretation tasks. It not only exhibits faster image generation speed but also enhances segmentation accuracy when contrasted with other GAN models like CoGAN, SimGAN, and CycleGAN~\cite{lv2021remote}.

However, it's worth noting that these advanced oversampling techniques come with their own set of challenges. One notable concern is the potential for overfitting the oversampled minority class. This risk primarily arises from the biases that can persist in the data even after applying these oversampling techniques~\cite{sampath2021survey}.

\subsubsection{Model-level approaches}

\textbf{Cost-sensitive learning} \\
Cost-sensitive learning involves considering the different costs associated with classifying data points into various categories. Instead of treating all misclassifications equally, it takes into account the consequences of different types of errors. For example, it recognises that misclassifying a rare positive instance as negative (more prevalent) is generally more costly than the reverse scenario.
In cost-sensitive learning, the goal is to minimise both the total cost resulting from incorrect classifications and the number of expensive errors. This approach helps prioritise the accurate identification of important cases, such as rare positive instances, in situations where the class imbalance is a concern~\cite{elkan2001foundations}.

Cost-sensitive learning finds application in spatial modelling, scenarios involving imbalanced datasets, or situations where the impact of misclassification varies among different classes or regions. Several studies have shown it is effective in this context~\cite{tsai2009forecasting,kang2022random,wu2023multi}.

\textbf{Boosting} 

Boosting algorithms are commonly used in geospatial modelling because they are superior in handling tabular spatial data and addressing imbalanced data~\cite{tien2016gis, song2018landslide, yu2020improving, kang2022random, kozlovskaia2017deep}. They effectively manage both bias and variance in ensemble models.

Ensemble methods such as Bagging or Random Forest reduce variance by constructing independent decision trees, thus reducing the error that emerges from the uncertainty of a single model. In contrast, AdaBoost and gradient boosting train models consecutively and aim to reduce errors in existing ensembles. AdaBoost gives each sample a weight based on its significance and, therefore, assigns higher weights to samples that tend to be misclassified, effectively resembling resampling techniques. 

In cost-sensitive boosting, the AdaBoost approach is modified to account for varying costs associated with different types of errors. Rather than solely aiming to minimise errors, the focus shifts to minimising a weighted combination of these costs. Each type of error is assigned a specific weight, reflecting its importance in the context of the problem. By assigning higher weights to errors that are more costly, the boosting algorithm is guided to prioritise reducing those particular errors, resulting in a model that is more sensitive to the associated costs~\cite{sun2009classification}.

This modification results in three cost-sensitive boosting algorithms: AdaC1, AdaC2, and AdaC3. After each round of boosting, the weight update parameter 
is recalculated, incorporating the cost items into the process~\cite{sun2007cost, sun2005parameter}.
In cost-sensitive AdaBoost techniques, the weight of False Negative is increased more than that of False Positive. AdaC2 and AdaCost methods can, however, decrease the weight of True Positive more than that of True Negative. Among these methods, AdaC2 was found to be superior for its sensitivity to cost settings and better generalisation performance with respect to the minor class~\cite{sun2009classification}.

\subsubsection{Combining model-level and data-level approaches} 
Modifications of the discussed techniques could be used as well. For instance, several techniques combine boosting, and SMOTE approaches to address imbalanced data. One such method is SMOTEBoost, which synthesises samples from the underrepresented class using SMOTE and integrates it with boosting. By increasing the representation of the minority class, SMOTEBoost helps the classifier learn better decision boundaries and boosting emphasises the significance of minority class samples for correct classification~\cite{chawla2002smote, cui2014improvement, kozlovskaia2017deep}. As for limitations, SMOTE is a complex and time-consuming data sampling method. Therefore, SMOTEBoost exacerbates this issue as boosting involves training an ensemble of models, resulting in extended training times for multiple models. Another approach is RUSBoost, which combines RUS (Random Under-Sampling) with boosting. It reduces the time needed to build a model, which is crucial when ensembling is the case, and mitigates the information loss issue associated with RUS~\cite{seiffert2009rusboost}. Thus, the data that might be lost during one boosting iteration will probably be present when training models in the following iterations.

Despite being a common practice to address the class imbalance, creating ad-hoc synthetic instances of the minority class has some drawbacks. 
For instance, in high-dimensional feature spaces with complex class boundaries, calculating distances to find nearest neighbours and performing interpolation can be challenging~\cite{chawla2002smote, he2009learning}. To tackle data imbalances in classification, generative algorithms can be beneficial. For instance, a framework combining generative adversarial networks and domain-specific fine-tuning of CNN-based models has been proposed for categorising disasters using a series of synthesised, heterogeneous disaster images ~\cite{eltehewy2023efficient}. 
SA-CGAN (Synthetic Augmentation with Conditional Generative Adversarial Networks) employs conditional generative adversarial networks (CGANs) with self-attention techniques to create high-quality synthetic samples~\cite{dong2022sa}. 
By training a CGAN with self-attention modules, SA-CGAN creates synthetic samples that closely resemble the distribution of the minority class, successfully capturing long-range interactions.
Another variation of GANs, EID-GANs (Extremely Imbalanced Data Augmentation Generative Adversarial Nets), focus on severely imbalanced data augmentation and employ conditional Wasserstein GANs with an auxiliary classifier loss~\cite{li2022eid}.

\section{Autocorrelation}


\subsection{Problem statement}


Autocorrelation is a statistical phenomenon where the value at a data point is influenced by the values at its neighbouring data points. In the context of environmental research, autocorrelation is frequently observed resulting from the spatial continuity of natural phenomena, such as temperature, precipitation, or species occurrence patterns.
However, the data-driven approaches applied for the tasks of spatial predictions assume independence among observations. If spatial autocorrelation (SAC) is not properly addressed, the geospatial analysis may result in misleading conclusions and erroneous inferences. Consequently, the significance of research findings may be overestimated, potentially affecting the validity and reliability of predictions \cite{schratz2019hyperparameter, salazar2022fair}. 

On the contrary, there could be environment-related tasks where autocorrelation is explored as the interdependence pattern between spatially distributed data not to be mitigated. For instance, based on an assessment of SAC catching regional spatial patterns in the LULC changes, a decision-support framework considering both land protection schemes, adapted financial investment and greenway construction projects supporting habitats was developed  \cite{li2022spatial}. Other examples are the enhancement of a landslide early warning system introducing susceptibility-related areas based on catching autocorrelation of landslide locations with rainfall variables \cite{tiranti2019shallow}, and an approach to assessing the spatiotemporal variations of vegetation productivity based on the SAC indices valuable for integrated ecosystem management \cite{ren2020measuring}.

\paragraph{Spatial autocorrelation}

\begin{figure}[!h]
\centering
\includegraphics[width=1\textwidth]{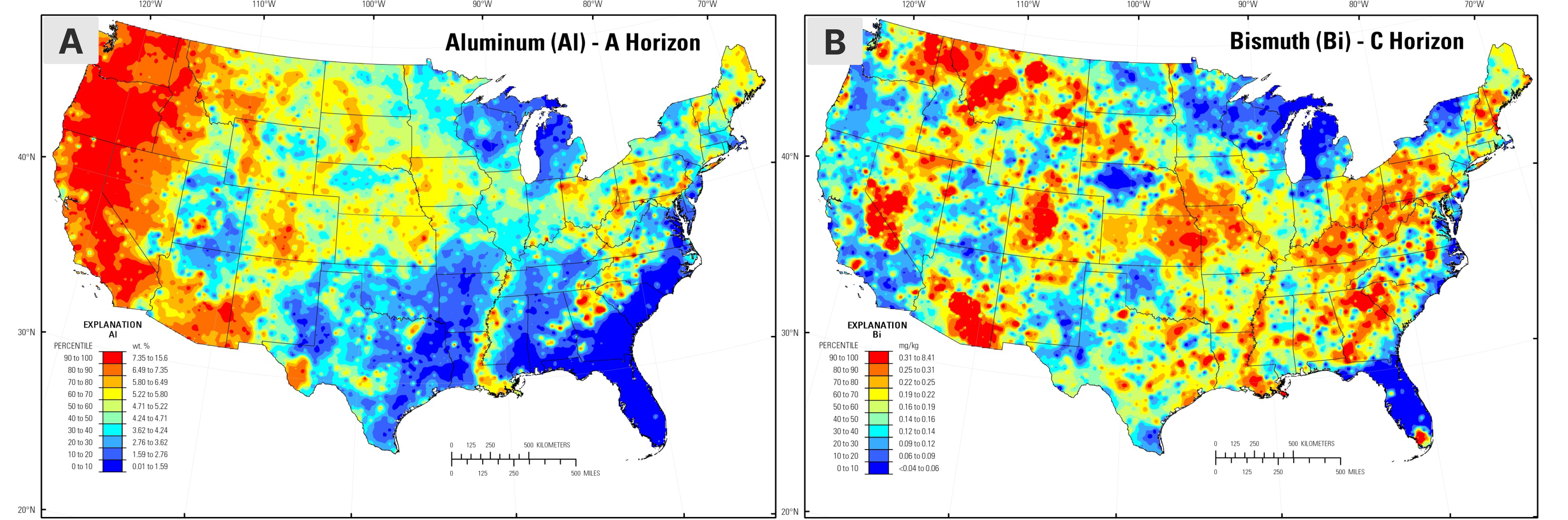}
\caption{The difference in spatial autocorrelation in Geochemical maps from USGS Open-File Report\cite{smith2013geochemical}. A) There appears to be a strong positive spatial autocorrelation with high concentrations (in red) and low concentrations (in blue) clustered together. B) The Bismuth map shows more scattered and less distinct clustering, indicating weaker spatial autocorrelation. The central and eastern regions show interspersed high and low values, suggesting a negative or weaker spatial autocorrelation.}
\label{fig:sac}
\end{figure}

While the definition of SAC varies, in general it integrates the principle that geographic elements are interlinked according to how close they are to one another, with the degree of connectivity fluctuating as a function of proximity, echoing the fundamental law of geography ~\cite{box1976time,hubert1981generalized}. Essentially, SAC outlines the extent of similarity among values of a characteristic at diverse spatial locations, providing a foundation for recognising and interpreting patterns and connections throughout different geographic areas \ref{fig:sac}.

Spatial processes exhibit characteristics of spatial dependence and spatial heterogeneity, each bearing significant implications for spatial analysis:
\begin{itemize}
    \item Spatial dependence. This phenomenon denotes the autocorrelation amidst observations, which contradicts the conventional assumption of residual independence seen in methods such as linear regression. One approach to circumvent this is through spatial regression.
    \item  Spatial heterogeneity. Arising from non-stationarity in the processes generating the observed variable, spatial heterogeneity undermines the effectiveness of constant linear regression coefficients. Geographically weighted regression offers a solution to this issue~\cite{leung2000testing,cho2010geographically}.
\end{itemize}

Numerous studies have ventured into exploring SAC and its mitigation strategies in spatial modelling. There exists a consensus that spatially explicit models supersede non-spatial counterparts in most scenarios by considering spatial dependence~\cite{gaspard2019residual}. However, the mechanisms driving these disparities in model performance and the conditions that exacerbate them warrant further exploration~\cite{crase2014incorporating, kim2016predicting, ching2019impact, ceci2019spatial}. A segment of the academic community contests the incorporation of autocorrelation in mapping, attributing potential positive bias in estimates as a consequence and advocating its application only for significantly clustered data~\cite{wadoux2021spatial}.


Residual spatial autocorrelation (rSAC) manifests itself not only in original data but also in the residuals of a model. Residuals quantify the deviation between observed and predicted values within the modelling spectrum. Consequently, rSAC evaluates the spatial autocorrelation present in the variance that the explanatory variables fail to account for. Grasping the distribution of residuals is vital in regression modelling, given that it underpins assumptions such as linearity, normality, equal variance (homoscedasticity), and independence, all of which hinge on error behavior~\cite{gaspard2019residual}.

\subsection{Approaches to measuring the problem of spatial autocorrelation}

Logically, the first step is to determine whether SAC is likely to affect the planned analysis — that is, if the model residuals display SAC, before considering modelling techniques that account for geographical autocorrelation. Checking for SAC has become commonplace in geography and ecology~\cite{dale2014spatial,dormann2007methods}. Among the methods used are 1) Moran's correlogram, 2) Geary's correlogram, and 3) variogram (semi-variogram)~\cite{bachmaier2008variogram}. 
The main idea of checking SAC underlies the investigation and tests whether nearby locations tend to be more clustered than randomness alone~\cite{fortin2009spatial}.

Moran’s I and Geary’s C are measures used to analyze spatial autocorrelation in data\cite{isaaks1989applied}. Moran’s I, ranging from -1 to +1, identifies general patterns within the entire dataset: values near +1 indicate clusters of similar values, -1 suggests adjacent dissimilar values, and 0 represents a random pattern. In contrast, Geary’s C, ranging from 0 to +2, is sensitive to local variations, with 0 indicating positive, 2 showing negative autocorrelation, and 1 denoting a random pattern. While Moran’s I is preferred for analyzing global patterns, Geary’s C is useful for detecting local patterns \cite{getis2007reflections}.

Correlograms based on  Moran's I typically exhibit a decline from a certain level of SAC to a value of 0 or even lower, signifying an absence of SAC at specific distances between locations. Essentially, a value of 0 or below suggests no observable SAC or a random spatial distribution of the variable under consideration. Similarly, for Geary's C, a value near 0 indicates an absence of SAC or spatial randomness, suggesting that the spatial distribution of the variable is akin to what might be expected if it were randomly distributed. On the other hand, higher values of Geary’s C, especially those greater than 1, suggest positive SAC. This means that the variable’s distribution shows similarity or clustering at different locations, highlighting a distinct spatial pattern in the data \cite{dormann2007methods}.


One of the crucial mathematical tools to assess the spatial variability and dependence of a stochastic variable is a variogram. Its primary purpose is to measure how the values of a variable alter as the spatial separation between sampled locations increases.

The variogram is mathematically defined as one-half of the variance of the dissimilarities observed between pairs of random variables at distinct locations, expressed as a function of the spatial separation between those locations.
In precise terms, the variogram represents the variance of the difference between the values of the spatial variable at two points, which are separated by the vector. In simpler terms, it quantifies the extent of dissimilarity or variation between pairs of observations at different spatial distances. The shape of the variogram cloud brings valuable insights into the spatial structure of the studied variable. Commonly employed variogram models, such as spherical, exponential, or Gaussian models, can be fitted to the scatter plot to estimate the parameters that characterize the spatial dependence.
The variogram holds significant importance in geostatistics and finds diverse applications, including spatial interpolation, prediction, and mapping of environmental variables such as soil properties, pollutant concentrations, and geological features. By comprehending the spatial structure through variogram analysis, researchers and practitioners can make more informed decisions and accurate predictions in fields such as geology, hydrology, environmental science, and related disciplines.

\subsection{Solutions to overcome SAC and rSAC}

The most common ways to eliminate the influence of SAC in the data on the prediction quality are the following:
\begin{enumerate}
\item proper sampling design
\item careful feature selection method
\item model selection
\item spatial cross-validation
\end{enumerate}

\subsubsection{Sampling design}

SAC influences occur in its capacity to delineate significance levels, demarcate discernible disparities in attribute measures across diverse populations, and elucidate attribute variability \cite{arbia1998error}. An amplified presence of SAC in georeferenced datasets invariably leads to an augmentation in redundant or duplicate information \cite{griffith2005effective}. This redundancy stems from two primary sources: geographic patterns informed by shared variables or the consequences of spatial interactions, typically characterised as geographic diffusion.

Exploring the details of sampling in relation to SAC reveals many layers of understanding:

\begin{itemize}
    \item The employment of diverse stratification criteria elicits heterogeneous impacts upon the amplitude of SAC \cite{di2018design}.
    \item The soil sampling density and SAC critically influence the veracity of interpolation methodologies \cite{radovcaj2021effect}.
    \item Empirical findings suggest that sampling paradigms characterised by heterogeneous sampling intervals — notably random and systematic-cluster designs — demonstrate enhanced efficacy in discerning spatial structures, compared with purely systematic approaches \cite{fortin1990spatial}.
\end{itemize}

The size of the sample also plays a key role in spatial modeling. In quantitative studies, it affects how broadly the results can be applied and how the data can be handled. In qualitative studies, it's crucial to establish that results can be applied in other contexts and for discovering new insights \cite{griffith2013establishing}. The relationship between SAC and the best sample size in quantitative research has been a popular topic, leading to many studies and discussions \cite{scott1993properties, griffith2005effective, dutilleul2011tests}.

In remote sensing, the main goal is often to use spectral data to guess attributes of places that have not been sampled. Regular sampling methods are usually best for this. Using close pairs of points in a regular design may make our predictions more accurate. But these designs do not work as well in different situations. Spatially detailed models are good for places with clear spatial patterns. They do not adapt well, however, to places with different patterns. Importantly, if our sampling design creates distances that match the natural spacing in the area, our predictions might be less certain \cite{rocha2020role}.

\subsubsection{Variable selection}
\label{variable_sel}

Spatial autocorrelation can be influenced significantly by selecting and treating variables within a dataset. Several traditional methodologies, encompassing feature engineering, mitigation of multicollinearity, and spatial data preprocessing, present viable avenues to address SAC-related challenges.

One notable complication arises from multicollinearity amongst the selected variables, which can potentiate SAC \cite{o2007caution}. Indications of multicollinearity are discernable through various diagnostic tools such as correlation matrices and variance inflation factors. To counteract multicollinearity, strategies encompassing the elimination of variables with high correlations and the application of dimensionality reduction techniques such as principal component analysis (PCA) can be employed. A judicious selection of pertinent variables, complemented by the development of novel variables hinged on domain expertise and exploratory data analysis, may further attenuate the manifestation of SAC.
Another approach for addressing this challenge is the consideration of rSAC across diverse variable subsets, followed by the deployment of classical model selection criteria like the Akaike information criterion \cite{cavanaugh2019akaike}. It is, however, imperative to recognise that the Akaike information criterion retains its efficacy in the context of rSAC when the independent variables do not exhibit spatial autocorrelation \cite{le2014spatial}.


In ML and DL, emerging methodologies have embraced spatial autocorrelation as an integral component. For instance, while curating datasets for training Long Short-Term Memory (LSTM) networks, an optimal SAC variable was identified and integrated into the dataset \cite{zhao2023hybrid}. Furthermore, spatial features, namely spatial lag and eigenvector spatial filtering (ESF), have been introduced to the models to account for spatial autocorrelation \cite{liu2022incorporating}. 

A novel set of features, termed the Euclidean distance field (EDF), has been innovatively designed based on the spatial distance between query points and observed boreholes. This design aims to seamlessly weave spatial autocorrelation into the fabric of ML models, further underscoring the significance of variable selection in spatial studies \cite{kim2023spatial}.

\subsubsection{Model selection}


Selecting or enhancing models to mitigate SAC impact is crucial. Spatial autoregressive models (SAR), especially simultaneous autoregressive models, are effective in this regard~\cite{anselin1988spatial}. SAR may stand for either spatial autoregressive or simultaneous autoregressive models. Regardless of terminology, SAR models allow spatial lags of the dependent variable, spatial lags of the independent variables, and spatial autoregressive errors. Spatial errors model (SEM), incorporate spatial dependence either directly or through error terms. SEMs handle SAC with geographically correlated errors. 
 Other approaches include auto-Gaussian models for fine-scale SAC consideration~\cite{lichstein2002spatial}. Spatial Durbin models further improve upon these by considering both direct and indirect spatial effects on dependent variables~\cite{lesage2008introduction}. Additionally, Geographically Weighted Regression (GWR) offers localised regression, estimating coefficients at each location based on nearby data~\cite{brunsdon1998geographically}.
In the context of SDM, six statistical methodologies were described to account for SAC in model residuals for both presence/absence (binary response) and species abundance data (Poisson or normally distributed response)~\cite{dormann2007methods}. These methodologies include autocovariate regression, spatial eigenvector mapping, generalised least squares (GLS), (conditional and simultaneous) autoregressive models, and generalised estimating equations. Spatial eigenvector mapping creates spatially correlated eigenvectors to capture and adjust for spatial autocorrelation effects~\cite{legendre2010comparison}. GLS extends ordinary least squares by considering a variance-covariance matrix to address spatial dependence~\cite{diniz2003spatial}.
The use of spatial Bayesian methods has grown in favour of overcoming SAC. Bayesian Spatial Autoregressive (BSAR) models and Bayesian Spatial Error (BSEM) models explicitly account for SAC by incorporating a spatial dependency term and a spatially structured error term, respectively, to capture indirect spatial effects and unexplained spatial variation~\cite{banerjee2014hierarchical}.
In recent years, the popularity of autoregressive models for spatial modelling as a core method has slightly decreased, while classical ML and DL methods have been extensively employed for spatial modelling tasks. Consequently, various techniques have been developed to leverage SAC’s influence effectively. The common approach is to incorporate SAC with the usage of autoregressive models during the stages of dataset preparation and variable selection. This approach is presented in greater detail in the previous subsection \ref{variable_sel}. On the other hand, combining geostatistical methods with ML is gaining popularity. 
For example, the usage of an artificial neural network (ANN) and the subsequent modelling of the residuals by geostatistical methods to simulate a nonlinear large-scale trend \cite{sergeev2019combining}.

\subsubsection{Spatial cross-validation}

Spatial cross-validation is a widely-used technique to account for SAC in various research studies~\cite{pohjankukka2017estimating,schratz2019hyperparameter,mila2022nearest,koldasbayeva2022large}. Neglecting the consideration of SAC for spatial data can introduce an optimistic bias in the results. This issue has been highlighted in the research, emphasising the importance of accounting for spatial dependence to obtain more accurate and unbiased assessments of model performance~\cite{fotheringham1999local, roberts2017cross, fortin2009spatial, negret2020effects}. For instance, it was shown~\cite{pohjankukka2017estimating} that random cross-validation could yield estimates up to 40 percent more optimistic than spatial cross-validation.

The main idea of spatial cross-validation is to split the data into blocks around central points of the dependence structure in space in space~\cite{roberts2017cross}. This ensures that the validation folds are statistically independent of the training data used to build a model. By geographically separating validation locations from calibration points, spatial cross-validation techniques effectively achieve this independence~\cite{zurell2020standard}.

Various methods are commonly employed in spatial cross-validation, including buffering, spatial partitioning, environmental blocking, or combinations thereof~\cite{roberts2017cross, ploton2020spatial}. These techniques aim to strike a balance between minimising SAC and avoiding excessive extrapolation, which can significantly impact model performance~\cite{roberts2017cross}. Buffering involves defining a distance-based radius around each validation point, excluding observations within this radius from model calibration. Environmental blocking groups data into sets with similar environmental conditions or clusters spatial coordinates based on input covariates~\cite{valavi2018blockcv}. Spatial partitioning, known as spatial K-fold cross-validation, divides the geographic space into K spatially distinct subsets through spatial clustering or using a coarse grid with K cells~\cite{roberts2017cross}.

However, it's worth mentioning an alternative discussion~\cite{wadoux2021spatial} showing that both standard and spatial cross-validation procedures may not be considered unbiased solutions for estimating the accuracy of mapping results, while the very concept of spatial cross-validation is heavily criticised. According to the results, neither standard nor spatial cross-validation provided satisfying results: map accuracy was overestimated for clustered data in the case of standard cross-validation or severely underestimated in the case of chosen spatial cross-validation strategies. Instead, probability sampling and design-based inference are suggested to obtain unbiased estimates of map accuracy in large-scale studies. Another concern is the request for better articulation of the meaning of validating a mapping model while examples of model validation and validation of the map are discussed.

In summary, spatial cross-validation techniques could be suitable to address SAC in data-based spatial modelling tasks while providing a transparent and precise description of the methodology of the model accuracy assessment and inference obtaining in a step-by-step manner is of high importance. Selecting the most suitable technique and its corresponding parameters should result from thoughtful consideration of the specificity of the research problem and the corresponding dataset.



\section{Uncertainty quantification}

\subsection{Problem statement}

\label{uq_section}
\import{}{uq_mikhail.tex}

\section{Practical tools}

In summary, it is crucial to consider the specificity of environmental data and its outcomes when selecting appropriate approaches for analysis. Table \ref{tab:tools} presents various methods to address the challenges discussed, including packages and libraries for geospatial analysis. Considering the dominant spread of Python and R as programming environments for data-based geospatial modelling, most of the tools to be implemented within these languages are provided. It is worth mentioning that although discussing libraries and packages are widely used in both academia and industry, common ML tools available in Python and R cover most of their functionality, being a replacement of these more specialized instruments with little change in quality and utility if used by advanced data scientists. 


        \begin{longtable}{c c c c}
        \caption{Geospatial data science tools in selected programming environments}
        \\
            \hline
            \multirow{1}{*}{\textbf{Solutions}} & \multicolumn{1}{l}{\textbf{Environment}} & \multicolumn{1}{l}{\textbf{Package/library}}& \multicolumn{1}{l}{\textbf{Description}} \\\cline{2-4}\hline
            \endfirsthead %
            
            \multirow{7}{*}{\parbox{3cm}{\centering Geospatial data analysis: general tools}} 
            & \multicolumn{1}{l}{R} & \multicolumn{1}{l}{sp \cite{bivand2008applied}}& \multicolumn{1}{l}{\parbox{7cm}{\vspace{0.2cm}Reading and writing spatial data represented by points, lines, polygons and grids, producing spatial objects, and performing spatial operations, e.g. plotting data as maps, spatial selection, retrieving coordinates, subsetting, print, summary.}} \\\cline{2-4}
            & \multicolumn{1}{l}{R} & \multicolumn{1}{l}{Simple Features (sf) \cite{pebesmasp2023}} & \multicolumn{1}{l}{\parbox{7cm}{\vspace{0.2cm}Reading, writing and converting Simple Features. Provides a set of tools for working with geospatial geometries represented by points, lines, polygons.}}\\\cline{2-4}
            & \multicolumn{1}{l}{R} & \multicolumn{1}{l}{raster \cite{hijmans2015package}} & \multicolumn{1}{l}{\parbox{7cm}{\vspace{0.2cm}Creating, reading, manipulating, and writing raster data. The package can process very large datasets. Raster algebra functions and high-level methods such as cropping and resampling are implemented.}}\\\cline{2-4}
            & \multicolumn{1}{l}{Python} & \multicolumn{1}{l}{geopandas\cite{kelsey_jordahl_2020_3946761}} 
            & \multicolumn{1}{l}{\parbox{7cm}{\vspace{0.2cm}Extends the datatypes used by python pandas to allow spatial operations on geospatial vector data. Reading, writing files, making maps and plots, data analysis and manipulations, such as buffering, intersection, and spatial joins.}}\\\cline{2-4}
            & \multicolumn{1}{l}{Python} & \multicolumn{1}{l}{rasterio\cite{gillies2023rasterio}}
            & \multicolumn{1}{l}{\parbox{7cm}{\vspace{0.2cm}Reading and writing gridded raster datasets, raster values can be extracted at precise points, and raster data can be warped and reprojected.}}\\\cline{2-4}
            & \multicolumn{1}{l}{Python} & \multicolumn{1}{l}{gdal\cite{gdal2023}} 
            & \multicolumn{1}{l}{\parbox{7cm}{\vspace{0.2cm} Geospatial Data Abstraction Library for raster and vector data reading, creating, writing, transformation and analysis}}\\\cline{2-4}
            & \multicolumn{1}{l}{Python} & \multicolumn{1}{l}{pysal\cite{pysal2007}} 
            & \multicolumn{1}{l}{\parbox{7cm}{\vspace{0.2cm}A family of packages for spatial data science: a collection of tools to explore, visualise and estimate relationships in spatial data with a focus on vector data. Among the core functions are detection of spatial clusters, hot-spots, and outliers, exploratory spatiotemporal data analysis, spatial regression and statistical modeling supported with inference obtaining.}} 
            \\\hline
            
            \multirow{2}{*}{\parbox{3cm}{\centering General oversampling methods of minority class}} 
            & \multicolumn{1}{l}{R} & \multicolumn{1}{l}{imbalance \cite{cordon2018imbalance}}& \multicolumn{1}{l}{\parbox{7cm}{\vspace{0.2cm}Set of tools to work with imbalanced datasets: oversampling algorithm based on the extension of the original SMOTE algorithm, filtering of instances and evaluation of synthetic instances, imbalance ratio computation.}} \\\cline{2-4}
            & \multicolumn{1}{l}{R} & \multicolumn{1}{l}{smotefamily\cite{smotefamily}} & \multicolumn{1}{l}{\parbox{7cm}{\vspace{0.2cm}A collection of various oversampling techniques for the minority class developed from SMOTE, including ADASYN, Borderline-SMOTE, DBSMOTE.}}
            \\\hline

            \multirow{2}{*}{\parbox{3cm}{\centering General oversampling methods of minority class and undersampling of the majority class}} 
            & \multicolumn{1}{l}{R} & \multicolumn{1}{l}{themis\cite{themis2023}}& \multicolumn{1}{l}{\parbox{7cm}{\vspace{0.2cm}Collection of techniques for balancing the data, including variations of SMOTE, ADASYN, ROSE, balancing majority class in a distance-based manned.}} \\\cline{2-4}
            & \multicolumn{1}{l}{Python} & \multicolumn{1}{l}{Imbalanced-learn\cite{JMLR:v18:16-365}} & \multicolumn{1}{l}{\parbox{7cm}{\vspace{0.2cm}Over-sampling methods include SMOTE, SMOTENC, SMOTEN, ADASYN, BorderlineSMOTE, KMeansSMOTE, SVMSMOTE. Under-sampling techniques include random under-sampling, algorithms based on cluster centroid of a KMeans algorithm, instance hardness threshold, variations of the nearest neighbour method, etc.}}
            \\\hline

            \multirow{1}{*}{\parbox{3cm}{\centering Spatial oversampling methods of minority class}} & \multicolumn{1}{l}{R} & \multicolumn{1}{l}{biomod2\cite{thuiller2016package}}& \multicolumn{1}{l}{\parbox{7cm}{\vspace{0.2cm}Ensemble Platform for Species Distribution Modeling, allows sampling minority class in a random, disk (select the locations of minority class based on the distance to prevalent class) and condition-based (a condition that differs from a defined proportion of prevalent class, could lead to over-optimistic result) manners.}} \\\hline

            \multirow{1}{*}{\parbox{3cm}{\centering Spatial undersampling of the majority class}} & \multicolumn{1}{l}{R} & \multicolumn{1}{l}{SpThin\cite{aiello2015spthin} }& \multicolumn{1}{l}{\parbox{7cm}{\vspace{0.2cm}SpThin provides spatial thinning of species occurrence records but can be applied for any other point-based spatial data. It is helpful for addressing problems associated with spatial sampling biases.}} \\\hline     

            \multirow{3}{*}{\parbox{3cm}{\centering Measuring SAC}} 
            & \multicolumn{1}{l}{R} & \multicolumn{1}{l}{spdep\cite{rpackages2022}}& \multicolumn{1}{l}{\parbox{7cm}{\vspace{0.2cm} Conducting spatial autocorrelation analysis, constructing spatial weight matrices, and visualizing spatial dependence patterns. Includes global and local Morans I and Gearys C, Hubert/Mantel general cross product statistic, Empirical Bayes estimates.}} \\\cline{2-4}
            & \multicolumn{1}{l}{R} & \multicolumn{1}{l}{ncf\cite{bjornstad2016package}} & \multicolumn{1}{l}{\parbox{7cm}{\vspace{0.2cm} Spatial analysis tool which is addressed to spatial autocorrelation, modelling semi-variograms, computing spatial covariance functions, and performing geostatistical interpolation.}}\\\cline{2-4}
            & \multicolumn{1}{l}{Python} & \multicolumn{1}{l}{esda\cite{pysal2007}} & \multicolumn{1}{l}{\parbox{7cm}{\vspace{0.2cm} Exploratory Spatial Data Analysis: Geary, Moran, Silhouette statistics are available.
            
            }} \\\hline   

            \multirow{1}{*}{\parbox{3cm}{\centering Spatial cross-validation}} & \multicolumn{1}{l}{R} & \multicolumn{1}{l}{blockCV\cite{valavi2018blockcv}}& \multicolumn{1}{l}{\parbox{7cm}{\vspace{0.2cm}Toolbox for cross-validation in spatial modelling. Includes tools for generating train and test folds for k-fold and leave-one-out cross-validation, to measure spatial autocorrelation ranges in candidate covariates, interactive graphical capabilities for creating spatial blocks and exploring data folds.}} \\\hline

            \multirow{4}{*}{\parbox{3cm}{\centering Uncertainty estimation}} 
            & \multicolumn{1}{l}{R} & \multicolumn{1}{l}{inlabru\cite{bachl2019inlabru}}& \multicolumn{1}{l}{\parbox{7cm}{\vspace{0.2cm} Spatial and general latent Gaussian modelling using integrated nested Laplace approximation. A prediction method based on fast Monte Carlo sampling allows posterior prediction of general expressions of the latent variables. Provide access to Bayesian inference from spatial point process, spatial count, gridded, and georeferenced data.}} \\\cline{2-4}
            & \multicolumn{1}{l}{R} & \multicolumn{1}{l}{Vizumap\cite{Vizumap2023}} & \multicolumn{1}{l}{\parbox{7cm}{\vspace{0.2cm} R package for visualising uncertainty in spatial data creating bivariate maps, pixel maps, glyph maps, and exceedance probability maps.}}\\\cline{2-4}
            & \multicolumn{1}{l}{R} & \multicolumn{1}{l}{spup\cite{heuvelink2007probabilistic}} & \multicolumn{1}{l}{\parbox{7cm}{\vspace{0.2cm} The package for examining the uncertainty propagation for input data and model parameters via the environmental model onto model outputs. The functions include uncertainty model specification, stochastic simulation and uncertainty propagation using Monte Carlo techniques. Probability distributions describe uncertain variables.}} \\\cline{2-4}
            & \multicolumn{1}{l}{Python} & \multicolumn{1}{l}{Uncertainty Toolbox\cite{chung2021uncertainty}} & \multicolumn{1}{l}{\parbox{7cm}{\vspace{0.2cm} A python toolbox for predictive uncertainty quantification, calibration, metrics, and visualizations.}}\\\hline

            \multirow{5}{*}{\parbox{3cm}{\centering Spatial modelling}} 
            & \multicolumn{1}{l}{R} & \multicolumn{1}{l}{biomod2\cite{thuiller2016package}}& \multicolumn{1}{l}{\parbox{7cm}{\vspace{0.2cm} Functions for modelling, calibration and evaluation, an ensemble of models, ensemble forecasting and visualization. Models include Random Forest, Boosted Regression Trees, Support Vector Machines, Artificial Neural Networks and others.}} \\\cline{2-4}
            & \multicolumn{1}{l}{JavaScript/Python} & \multicolumn{1}{l}{Google Earth Engine\cite{gorelick2017google}} & \multicolumn{1}{l}{\parbox{7cm}{\vspace{0.2cm} Catalog of satellite imagery and geospatial datasets and collection of tools for data retrieving, geospatial analysis and modelling.}} \\\cline{2-4}
            & \multicolumn{1}{l}{R} & \multicolumn{1}{l}{sdmTMB\cite{sdmTMB2022}} & \multicolumn{1}{l}{\parbox{7cm}{\vspace{0.2cm} Implements spatial and spatiotemporal Generalized Linear Mixed Effect Models.}}\\\cline{2-4}
            & \multicolumn{1}{l}{Python} & \multicolumn{1}{l}{verde\cite{uieda2018}} & \multicolumn{1}{l}{\parbox{7cm}{\vspace{0.2cm}Provides classes and functions for processing spatial data, like bathymetry, GPS, temperature, gravity, or anything else that is measured along a surface. The main focus is on methods for gridding such data (interpolating on a regular grid).}} \\\cline{2-4}
            & \multicolumn{1}{l}{Python} & \multicolumn{1}{l}{GSTools\cite{muller2022gstools}} & \multicolumn{1}{l}{\parbox{7cm}{\vspace{0.2cm} Provides methods for generating random fields and performing simple, ordinary, universal and external drift kriging and variogram estimation. }}
            \\\hline  

        \end{longtable}
    \label{tab:tools}

\section{Key areas for focus and growth}

Geospatial modelling has grown rapidly, driven by data-based models and the integration of ML and DL alongside traditional geospatial statistics. Previous sections highlighted common implementation gaps and approaches to address them. Additionally, it is worth exploring and discussing future developments and key possibilities concerning challenges in data-driven geospatial modelling.
Below, we highlight the major points of growth that can lead to new seminal works in this area.

\paragraph{New generation of datasets} 



It is crucial to enhance data quality, quantity, and diversity to ensure reliable models. Establishing well-curated databases in environmental research is of utmost importance as it drives scientific progress and industrial innovation. When combined with modern tools, these databases can contribute to developing powerful models.

A particular area of interest is the collection of cost-effective and efficient semi-supervised data, which typically has limited labels. Although currently underdeveloped, this data type holds significant potential for expansion and improvement.
In computer vision and natural language processing, the superior quality of recently introduced models often comes from using more extensive and better datasets.
Internal Google dataset on semi-supervised data JFT-3B with nearly three billion labelled images led to major improvements~\cite{zhai2022scaling,sun2017revisiting}. Another major computer vision dataset example is LVD-142M with about 142 million images \cite{oquab2023dinov2}.
We note that the paper provides a pipeline that can be used to extend the size of existing datasets to two orders of magnitude.
In natural language processing, a recent important example is training large language models. 
\cite{touvron2023llama}.
It uses a preprocessed dataset with 2 trillion tokens.
More closely related to geospatial modelling is the adoption of climate data. It now also allows the application of DL models mainly due to the increasing number of available measurements.
For example, SEVIR dataset~\cite{veillette2020sevir} allowed better prediction via a variant of Transformer architecture~\cite{gao2022earthformer}.
In~\cite{ravuri2021skilful}, the authors developed a model for precipitation nowcasting. 
To train the model, they employ radar measurements at a grid with cells of $1 \times 1$ kilometres, taking every $5$ minutes for $3$ years. 
In total, around 1 TB of data were used.

Furthermore, integrating diverse data sources offers a promising path forward. Combining datasets from various domains, such as satellite imagery, meteorological and climatic data, and social data, such as social media posts that provide real-time environmental information for specific locations, can be beneficial. By developing multimodal models capable of processing these diverse data sources, the community can enhance model robustness and effectively address the challenges discussed in this study and the existing literature. Most of the research combines image and natural language modalities~\cite{zeng2022socratic}, while other options are possible. 

\paragraph{New generation of models} 

The continuous advancement of technology has led to the emergence of more sophisticated data sources, including higher-resolution remote sensing and more accurate geolocation data. Additionally, human efforts contribute to high-quality curated data. While this is beneficial, it presents challenges in adapting existing geospatial models to handle such data. Traditional models may need to be more suitable and efficient, necessitating the developing and validation of new models and computational methods. Incorporating DL methods is a potential solution, although they come with challenges related to interpretability and computational efficiency, especially when dealing with large volumes of data. We anticipate the emergence of self-supervised models trained on large semi-curated datasets for geospatial mapping in environmental research, similar to what we have seen in language modelling and computer vision. Such modelling approaches have also been applied to satellite images~\cite{mohanty2020deep} including, for example, a problem of the state of plants estimation~\cite{illarionova2022estimation} and assessment of damaged buildings in disaster-affected area~\cite{novikov2018satellite}.



\paragraph{Producing industry-quality solutions: deployment and maintenance.}
After constructing a model, it needs to be deployed in a production environment. 
Access to necessary data and supporting services is crucial to ensure safe and continuous operation. 
Another challenge is the ageing of data-based models caused by environmental factors like changing climate ~\cite{burnaev2022fundamental}, shifts in data sources, or transformations in output variables, e.g., alterations of land use and land cover~\cite{kenthapadi2022model}. Monitoring and considering such changes is essential to either discontinue using an outdated model or retrain it with new data~\cite{gama2014survey}. 
The monitoring schedule can vary, guided by planned validation checking or triggered by data corruption as well as new business process implementations. Deployment and maintenance are often underestimated despite requiring significant resources and additional steps for long-term success~\cite{vela2022temporal}. Another area of possible growth is related to developing new methods, including advanced DL methods. 
Incorporation of concept drift into the maintenance process is also an option~\cite{van2022three}.

\bibliography{sample}


\end{document}

%% file: uq_mikhail.tex
\begin{figure}[!h]
\centering
\includegraphics[width=\textwidth]{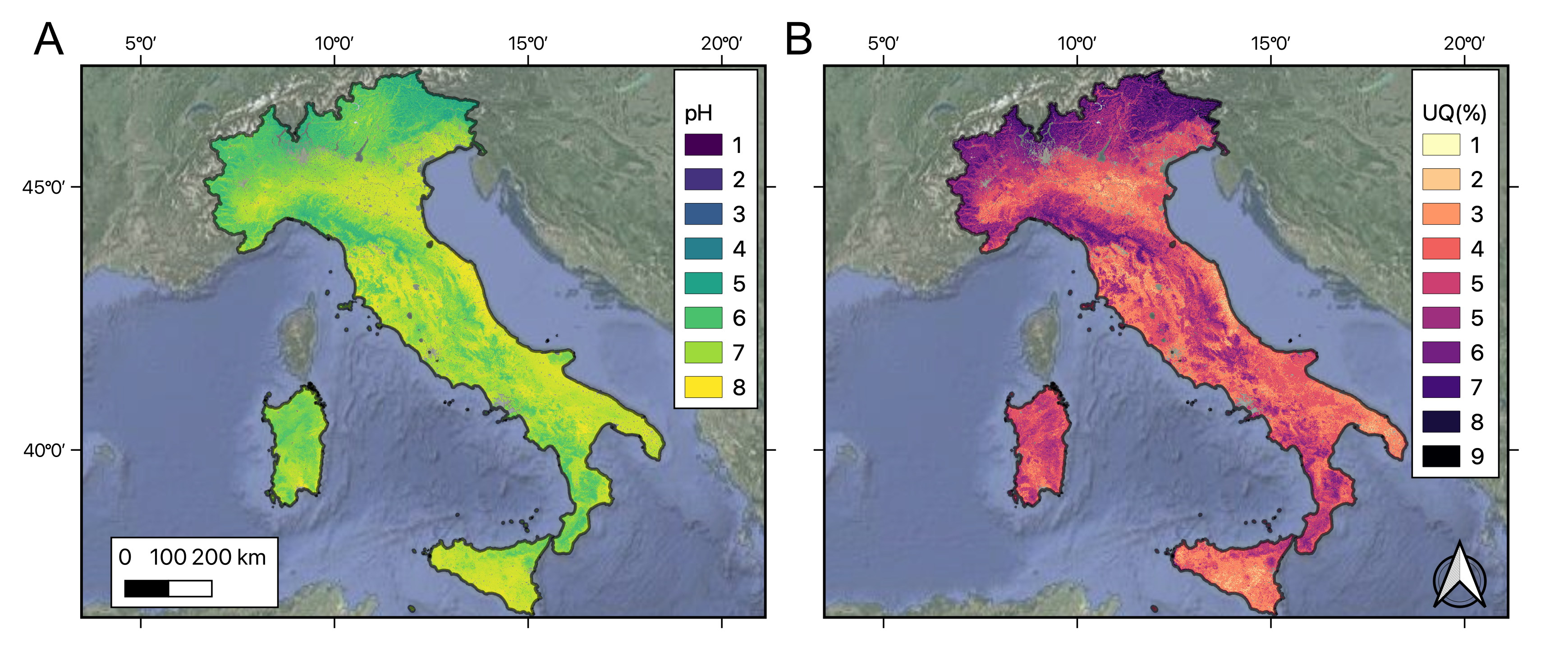}
\caption{Example of uncertainty quantification (UQ) for spatial mapping provided within the project SoilGrids \cite{poggio2021soilgrids} a) maps of one of the target variables -- soil pH(water) in the topsoil layer; b) maps of associated uncertainty estimated using prediction interval coverage probability (PICP) index for the same territory.}
\label{fig:uq_maps}
\end{figure}

Geospatial predictions using machine learning have become convenient for routine decision-making workflows. To ensure these predictions are reliable and sufficient, assessing the uncertainty associated with the model’s forecasts is crucial. Uncertainty quantifies the level of confidence the model has in its predictions (Figure \ref{fig:uq_maps}). Two primary types of uncertainty exist: aleatory uncertainty, which arises from data uncertainty, and epistemological uncertainty, which originates from knowledge limitations \cite{abdar2021review}.

Sources of uncertainty may stem from incomplete or inaccurate data, inaccurately specified models, inherent stochasticity in the simulated system, or gaps in our understanding of the underlying processes. Assessing aleatoric uncertainty caused by noise, low spatial or temporal resolution, or other factors which cannot be taken into account can be challenging. For that reason, the most of research is focused on epistemic uncertainty. Reducing uncertainty in ML models is essential for improving their reliability and accuracy.

\subsection{Solutions for uncertainty quantification} 
\subsubsection{Classical ML approaches}
One of the common approaches for uncertainty quantification (UQ) in geospatial modelling is quantile regression \cite{bassett1978asymptotic}. It allows one to understand not only the average relationship between variables but also how different quantiles (percentiles) of the dependent variable change with the independent variables. In other words, it helps to analyze how the data is distributed across the entire range, rather than just focusing on the central tendency. Quantile regression is particularly useful when dealing with data that may not follow a normal distribution or when there are outliers in the data that could heavily influence the results.  




For instance, to quantify the uncertainty of models for nitrate pollution of groundwater, quantile regression and uncertainty estimation based on local errors and clustering (UNEEC) \cite{shrestha2006machine} were used \cite{rahmati2019predicting}. Quantile regression was also used for the UQ of four conventional ML models for digital soil mapping: to estimate UQ the authors analysed mean prediction intervals (MPI) and prediction interval coverage probability (PICP) \cite{kasraei2021quantile}. Another widely used technique for UQ is bootstrap, which is a statistical resampling technique that involves creating multiple samples from the original data to estimate the uncertainty of a statistical measure \cite{efron1992bootstrap, heskes1996practical}. One more metric is mean-variance estimation (MVE), which is used to simultaneously estimate both the mean (average) and variance (spread) of a dataset. It helps to describe the central tendency and variability of the data \cite{nix1994estimating}.


\subsubsection{Gaussian process regression}



Gaussian Process Regression, also known as kriging, is commonly used for UQ in geospatial applications, providing a natural way to estimate the uncertainty associated with spatial predictions. In a study focused on spatiotemporal modelling of soil moisture content using neural networks, the authors utilised sequential Gaussian simulations to estimate uncertainty and reduced RMSE by 18\% in comparison with the classical approach \cite{song2016modeling}. 
Another approach, known as Lower Upper Bound Estimation, was applied to estimate sediment load prediction intervals generated by neural networks \cite{chen2019uncertainty}. 
For soil organic mapping, researchers compared different methods, including sequential Gaussian simulation (SGS), quantile regression forest (QRF), universal kriging, and kriging coupled with random forest. They concluded that SGS and QRF provide better uncertainty models based on accuracy plots and G-statistics \cite{szatmari2019comparison}. However, Random Forest demonstrated better performance of prediction uncertainty in comparison with kriging in soil mapping in another study \cite{takoutsing2022comparing}, although predictions of regression kriging were found to be more accurate, that can be related to the architecture of these models.

\subsubsection{Bayesian techniques}

Another approach to estimating uncertainty in ML models is through Bayesian inference \cite{ellison2004bayesian}. In Bayesian methods, model parameters are treated as random variables with prior distributions, allowing for uncertainty modelling. However, with its complex relationships and spatial dependencies, geospatial modelling poses challenges in uncertainty quantification. To estimate uncertainty in model predictions, the posterior distribution of parameters given the data and priors is used.

Bayesian techniques have been applied to various models, including neural networks, Gaussian processes, and spatial autoregressive models, to estimate uncertainty in predictions of variables such as temperature, air quality, and land use. Some main methods for uncertainty quantification using Bayesian neural networks include Monte Carlo (MC) dropout \cite{gal2016dropout}, sampling via Markov chain Monte Carlo (MCMC) \cite{kupinski2003ideal}, and Variational autoencoders \cite{swiatkowski2020k}. However, it should be noted that most of these methods are specifically used for uncertainty quantification in DL and at the moment they are not widely implemented in geospatial modeling \cite{abdar2021review}. 

For instance, Bayesian techniques have been used in weather modelling, particularly wind speed prediction and hydrogeological calculations, to analyse the risk of reservoir flooding \cite{lu2020risk}. Probabilistic modelling was employed to assess the uncertainty of spatial-temporal wind speed forecasting, with models based on spatial-temporal neural networks using convolutional GRU and 3D CNN. Variational Bayesian inference was also utilised \cite{liu2020probabilistic}. Similarly, Bayesian inference has been applied to estimate uncertainty in soil moisture modelling \cite{harrison2012quantifying}. Another study used Bayesian inference to model the spread of invasive species \cite{cook2007bayesian}.

\subsubsection{Ensemble techniques}
Model ensembling is a powerful technique used in geospatial modelling to address uncertainty. 
Geospatial models often deal with complex systems where uncertainty arises from various sources, including input data, parametrisation, and modelling assumptions. 
Ensembles can help both during the reduction and estimation of uncertainty.
The diversity of predictions from different members of an ensemble serves as a natural way to estimate uncertainty.
On the other hand, more robust and reliable estimates can be obtained by combining predictions from multiple models through ensembling methods like weighted averaging, stacking, or Bayesian model averaging. Ensembling helps mitigate uncertainties associated with individual models and provides a way to estimate uncertainty by computing the variance of predictions across the ensemble \cite{meinshausen2006quantile}. To solve the problems of spatial mappings, such as equifinality, uncertainty and conditional bias, ensemble modelling and bias correction framework were proposed. The method was developed for mapping soils using the XGBoost model and environmental covariances as predictors. It was shown that ensemble modelling helped solve the equifinality problem in the data set while demonstrating better performance \cite{sylvain2021using}. Another example is the comparison of regional and global ensemble models for soil mapping \cite{brungard2021regional}. It was found that the performance of an ensemble of regional models was the same as global models, but regional model ensembles had less uncertainty than global models. 
Ensembling approaches to UQ were also applied in the DL modelling tasks \cite{pearce2018high}. In another study, authors proposed a system that combines ML models within a spatial ensemble framework to reduce uncertainty and enhance the accuracy of site index predictions \cite{gavilan2021reducing}. For soil clay content mapping, authors estimated the uncertainty of seven ML models and their ensembles \cite{zhao2022clay}. Ensembling proves to be a valuable technique in geospatial modelling, as it leverages the collective knowledge of multiple models to improve predictions and provide more comprehensive uncertainty estimates.

\subsection{Solutions to address the uncertainty in spatial predictions}

Several approaches can be used to reduce uncertainty, falling into two groups. The first group is related to input data and involves increasing data quality, using more data from different domains, and adding feature engineering to select predictors highly relevant to the problem. This can help the model focus on the most informative aspects of the data, reducing uncertainty caused by irrelevant or redundant features. The second group is devoted to the modelling step and includes spatial and temporal cross-validation, model regularisation techniques to prevent overfitting, combining multiple models through techniques such as bagging or boosting, and more complex approaches such as Bayesian methods, Gaussian techniques, and transfer learning, which are described above.

Visualisation methods for UQ in geospatial modelling hold a distinct place compared to other areas of ML. Researchers emphasize the significance of visually analyzing maps with uncertainty estimates, especially for biodiversity and policy conversation tasks \cite{jansen2022stop}. Visualisation techniques such as bivariate choropleth maps, map pixelation, and glyph rotation to represent spatial predictions with uncertainty can be used \cite{lucchesi2017visualizing}.